\documentclass[conference]{IEEEtran}

\usepackage{cite}
\usepackage{amsmath,amssymb,amsfonts}
\usepackage{algorithmic}
\usepackage{graphicx}
\usepackage{textcomp}
\usepackage{xcolor}
\def\BibTeX{{\rm B\kern-.05em{\sc i\kern-.025em b}\kern-.08em
    T\kern-.1667em\lower.7ex\hbox{E}\kern-.125emX}}

\usepackage{bbm}
\usepackage{multirow}
\usepackage[ruled,linesnumbered]{algorithm2e}
\usepackage{booktabs}
    
\begin{document}

\title{MTS-DMAE: Dual-Masked Autoencoder for Unsupervised Multivariate Time Series Representation Learning}

\author{
\IEEEauthorblockN{Yi Xu, Yitian Zhang, Yun Fu}
\IEEEauthorblockA{Northeastern University, Boston, USA\\
xu.yi@northeastern.edu, markcheung9248@gmail.com, yunfu@ece.neu.edu}
}

\maketitle

\begin{abstract} 
Unsupervised multivariate time series (MTS) representation learning aims to extract compact and informative representations from raw sequences without relying on labels, enabling efficient transfer to diverse downstream tasks. In this paper, we propose Dual-Masked Autoencoder (DMAE), a novel masked time-series modeling framework for unsupervised MTS representation learning. DMAE formulates two complementary pretext tasks: (1) reconstructing masked values based on visible attributes, and (2) estimating latent representations of masked features, guided by a teacher encoder. To further improve representation quality, we introduce a feature-level alignment constraint that encourages the predicted latent representations to align with the teacher's outputs. By jointly optimizing these objectives, DMAE learns temporally coherent and semantically rich representations. Comprehensive evaluations across classification, regression, and forecasting tasks demonstrate that our approach achieves consistent and superior performance over competitive baselines.
\end{abstract}


\section{Introduction}
Multivariate time series (MTS) data constitute a fundamental modality that appears in numerous application areas.
Typically, MTS data describe the temporal evolution of multiple synchronized variables, such as simultaneous measurements of diverse physical quantities.
Despite the growing interest in MTS data, obtaining labeled data remains scarce because labeling such data is challenging and often requires specialized infrastructure or domain expertise~\cite{wu2022timesnet}. Consequently, there is a strong demand for methods that can deliver high performance while relying on limited labeled data or unlabeled data.

Unsupervised representation learning has been widely recognized as a simple yet powerful methodology that can extract meaningful representations from unlabeled data for enabling downstream tasks. 
Despite notable advances in fields like computer vision~\cite{bao2022beit} and natural language processing~\cite{chowdhery2022palm}, relatively little attention has been given to MTS data.
One key bottleneck is the lack of diverse forms of supervision, including attributes, contexts, and relationships, across multiple levels of granularity, which are more commonly available in other data types. 
Given the complex and unique nature of MTS data, exploring its inherent properties through unsupervised representation learning is essential for enhancing its representation capability.

Existing research on unsupervised MTS representation learning generally falls into two groups: (1) pretext task-based methods and (2) contrastive learning–based methods. 
In the first category, the goal is to train the model to excel in a specific pretext task, often intentionally more challenging than the eventual target task, to achieve effective generalization to the latter. 
While conceptually straightforward, the major challenge lies in defining a pretext task that can offer simultaneous benefits across multiple downstream tasks. Conversely, the second category leverages contrastive loss as a common strategy for representation learning. 
The fundamental concept involves designing diverse sampling strategies that incorporate appropriate augmentations to produce positive and negative sample pairs. 
However, such approaches are vulnerable to false negatives and face difficulties in modeling long-range dependencies, mainly due to the missing global context.

In this study, we focus on the first category by formulating pretext tasks for unsupervised multivariate time series representation learning. 
Motivated by these challenges, we propose a \underline{D}ual-\underline{M}asked \underline{A}uto-\underline{E}ncoder (DMAE) tailored for masked time-series modeling (MTM), where we introduce a pair of dual completion pretext tasks aimed at facilitating the acquisition of compact feature representations from MTS. 

The first task is to predict the masked attributes of MTS data via a transformer-based encoder, while the second task is to estimate the masked representations under the supervision of a teacher branch. First, a pair of complementary masks is generated, and the given MTS sample is masked by these two masks separately. Then, one masked sample is fed into the transformer-based encoder to extract the feature representations, and the other is fed into the teacher branch, which consists of one teacher encoder and one feature-query module. The teacher encoder shares the parameters with the aforementioned transformer-based encoder to generate the supervision signal, and the feature-query module is proposed to predict the masked feature representations conditioned on the visible feature representations. 
Specifically, the cross-attention mechanism is employed, where the initial queries are derived from the input to the teacher branch via a linear projection to output the predicted feature representations. 
Finally, the representations of masked features predicted by the model are enforced to align with those computed by the teacher encoder via our proposed alignment loss.
In addition, we also include a decoder to map the predicted masked feature representations to the MTS attributes to reconstruct the masked attributes. 
In comparison to previous MTM methods~\cite{zerveas2021transformer}, which couple the encoding process with pretext task completion, our method explicitly separates representation learning from task execution.
This separation strengthens the quality of learned representations and yields greater benefits for downstream tasks.
The main contributions of this work are:
\begin{itemize}
    \item We introduce a Dual-Masked Autoencoder (DMAE) for unsupervised representation learning of multivariate time series via masked modeling.
    \item We incorporate a teacher–student architecture with a latent alignment constraint, enabling semantic consistency and decoupling representation learning from the attribute completion objective.
    \item Extensive evaluations across classification, regression, and forecasting benchmarks demonstrate that DMAE consistently achieves strong performance and generalizes well across diverse datasets, while maintaining lightweight inference.
\end{itemize}

\section{Related Work}
\subsection{Unsupervised Representation Learning for Time Series}
Unsupervised representation learning has been extensively studied in computer vision and natural language processing, but remains relatively underexplored in the time series domain. 
Current methods for unsupervised sequential data learning can be grouped into reconstruction-based and contrastive learning–based paradigms.

The first category centers around designing pretext tasks that require the model to reconstruct missing or corrupted input values. Classical methods adopt autoencoders~\cite{malhotra2017timenet} or sequence-to-sequence networks~\cite{lyu2018improving} to reconstruct the original sequence, thus implicitly capturing temporal dependencies. For clustering tasks, several works~\cite{fortuinsom2018somvae,kopf2021mixture} propose learning distance-preserving representations by approximating pairwise similarity between sequences. For instance, Dynamic Time Warping (DTW)~\cite{lei2017similarity} is used to measure temporal alignment, while~\cite{ma2019learning} integrates reconstruction with K-means~\cite{krishna1999genetic} to produce cluster-specific temporal embeddings.
Recent advances adopt masked modeling strategies, where the model learns to recover randomly masked input variables. For example,~\cite{zerveas2021transformer,chowdhury2022tarnet,dong2023simmtm} apply masking schemes on multivariate time series and train the model to predict missing attributes. These approaches achieve strong performance, but mainly focus on reconstructing observable values, without explicitly constraining the latent representations. In contrast, our work introduces a dual-masked autoencoder that not only reconstructs masked inputs but also aligns the student’s latent features with those of a teacher network, thereby enhancing semantic consistency at the representation level.

The second category is based on contrastive learning, which relies on constructing positive and negative sample pairs through diverse augmentation techniques~\cite{yang2022unsupervised,yue2022ts2vec,luo2023time}. A representative example is~\cite{franceschi2019unsupervised}, which uses a triplet loss inspired by Word2Vec to learn representations via segment sampling. More recent methods, such as~\cite{eldele2021time,tonekaboni2021unsupervised,zhang2022self,wang2022learning,woo2021cost,tonekaboni2022decoupling,zhang2022dbt,zheng2023simts,liu2024focal}, propose increasingly fine-grained or domain-aware strategies for constructing contrastive pairs. Among them, SimTS~\cite{zheng2023simts} incorporates contrastive objectives into masked pretraining, while FOCAL~\cite{liu2024focal} enhances frame-level alignment for more precise encoding.

Although contrastive learning has proven effective, it suffers from limitations such as sensitivity to false negatives and reliance on carefully designed augmentations. Moreover, the contrastive loss is often applied at the global sequence level, potentially neglecting local temporal consistency. Our method avoids these issues by adopting a reconstruction-based learning objective, while introducing a feature-level alignment constraint that preserves temporal semantics in the latent space.

\begin{figure*}[ht]
  \centering
   \includegraphics[width=0.9\linewidth]{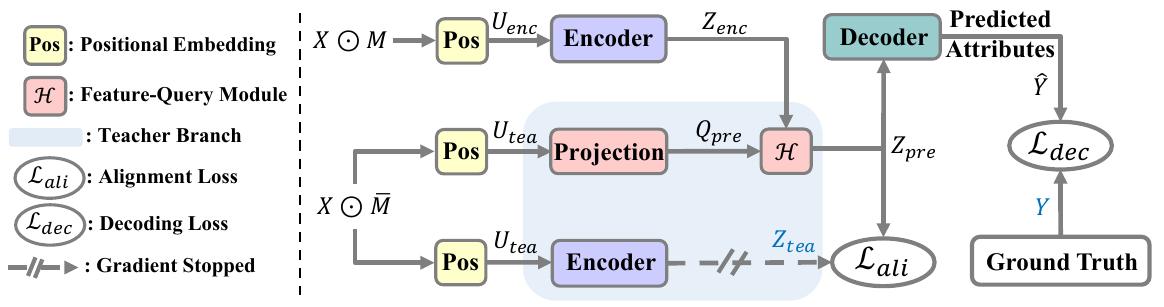}
   \caption{An overview of our proposed framework. Here, $X$ represents the multivariate time series sample, $M$ the mask, and $\odot$ the element-wise multiplication.}
   \label{fig:framework}
\end{figure*}

\subsection{Transformer for Time Series}
After Transformer~\cite{vaswani2017attention} was first presented, it immediately achieved significant success in enormous domains, and a large number of its variants have been developed in many applications, i.e., natural language processing~\cite{bao2022beit}, vision transformer~\cite{liu2021swin}. 
Given the great capacity for modeling long-range dependencies, the Transformer has been increasingly adopted in the time series domain.
For instance, many transformer-based methods~\cite{zhou2021informer,wu2021autoformer,zhou2022fedformer} have recently been developed for time series forecasting, achieving promising results compared with statistical and RNN-based methods.
Moreover, the self-attention mechanism has been further extended to other time series tasks, including classification~\cite{xie2022marina,ren2022autotransformer}, regression~\cite{perrier2022attention}, and forecasting~\cite{zhang2022crossformer,zeng2023transformers}.

Considering the unique advantage of modeling long-term dependencies, we adopt the transformer architecture as the encoder in our work. In contrast, we focus on building a broad framework to learn expressive enough representations that transfer effectively to diverse downstream tasks such as classification, regression, and forecasting, rather than optimizing solely for a single task.

\section{Preliminary}
Given its widespread adoption in handling temporal data, we employ the Transformer architecture as our method backbone. 
Transformers, along with their various adaptations, have gained promising performance in different domains.
By leveraging self-attention, Transformers can effectively process sequential inputs of varying lengths while enabling parallel computation across the sequence. The architecture comprises three main components: (1) the self-attention mechanism, (2) multi-head attention, and (3) position-wise feed-forward networks. Given the input sequence $X$, it is projected to high-dimensional matrices as follows:
\begin{equation}
    Q, K, V = linear(X),
    \label{eq:qkv}
\end{equation}
where $Q$, $K$, and $V$ represent the query, key, and value matrices, respectively. Then, the attention score is calculated as follows:
\begin{equation}
    Attention(X) = softmax(QK^T/\sqrt{d})V,
\end{equation}
where $d$ is the dimension of the keys.

To strengthen representational power, the Transformer adopts multi-head attention, where several self-attention mechanisms are applied in parallel and the outputs are concatenated through linear projections. This operation is defined as follows:
\begin{equation}
    \begin{aligned}
    MultiHead(X) & = Concat(head_1, ..., head_h)W^{O} \\
    head_i & = Attention(XW^Q_i, XW^K_i, XW^V_i)
    \end{aligned}
    .
\end{equation}

Additionally, two linear transformations with a $ReLU$ activation function are applied between multi-head attention blocks. The feed-forward network is defined as follows:
\begin{equation}
    FFN(X) = max(0, XW_1+b_1)W_2 + b_2.
\end{equation}

The Transformer applies Layer Normalization (LN) to enhance training stability and improve convergence speed.
However, as pointed out in~\cite{shen2020powernorm}, LN is not suitable for time series relevant tasks since there are many outliers and the noisy data will negatively influence the normalization effect. Therefore, we adopt Batch Normalization (BN) instead of LN in our implementation.

\section{Method}
Fig.~\ref{fig:framework} depicts the architecture of our method, which includes three essential components: (1) encoder, (2) teacher branch, and (3) decoder.
Our goal is to enable fully unsupervised learning of robust and semantically consistent representations for multivariate time series.
To this end, we propose a Dual-Masked Autoencoder framework that introduces the following three key innovations:

(1) \textbf{Dual-branch masking:} Inspired by denoising and masked modeling paradigms, we design two parallel masking streams—a student branch observing partial input and a teacher branch observing a complementary view. This dual-masked design encourages the model to reason from incomplete observations and enhances feature robustness.

(2) \textbf{Latent alignment with a teacher:} Beyond reconstructing missing values, we introduce an alignment constraint that encourages the student encoder’s hidden features to match those of the teacher. This ensures temporal consistency and smoothness in the learned latent space.

(3) \textbf{Multi-objective learning:} The student is jointly trained to perform two pretext tasks, attribute-level imputation and latent-level alignment, under a unified transformer-based architecture. Our model can be plugged into any time-series encoder structure without additional inference-time cost.

\subsection{Model Input and Masking Strategy}~\label{sec:modelinput}
The inputs to our model are MTS samples, where for each sample, we define it as $X \in\mathbb{R}^{M\times T}$ with $M$ different variables and sequence length $T$. At each time stamp, we denote $x_{t} \in\mathbb{R}^{M}$ as the attribute vector that constitutes the sequence $X$, i.e., $X=\left[x_{1},..., x_{t},..., x_{T}\right], t\in \left[1, T \right]$. 
We standardize the vector $x_{t}$ by subtracting the mean and dividing by the variance across the training set.

We have adopted a masking strategy for time series data, which differs from the strategy commonly used for images, i.e., dividing each image into 16 or 64 patches and randomly selecting a certain number of patches to mask. Our masking strategy operates at the attribute level of time series, and this is achieved by defining a binary mask $M$ for each time-series sample $X$, where $0$ represents the attribute is masked and $1$ represents the attribute is not masked. The binary mask $M\in\mathbb{R}^{M\times T}$ has the same shape as $X$ and is generated independently for each sample. The input is then obtained via element-wise multiplication between $X$ and $M$ as follows:
\begin{equation}
    \hat{X} = X \odot M.
\end{equation}

Instead of employing the Bernoulli distribution to generate masks, we apply a geometric distribution with mean $l_{m}$ to determine the masked length, and therefore the unmasked mean length $l_{u}$ is defined as follows:
\begin{equation}
    l_{u}= \frac{1-r}{r}l_{m},
\end{equation}
where we set $l_{m}=3$ and $r=0.15$ in our work. This masking strategy, which we refer to as span masking, is designed to increase the possibility of masking the sequence for a continuous period. The insight behind such design is that each attribute has an intrinsic temporal correlation with others, the continuous masking makes the pretext task more challenging.

\subsubsection{Input to the Encoder}
Given the masked time series $\hat{X}$, we first project $\hat{X}$ to the high dimensional feature matrix $\hat{X}_{\text{enc}}\in\mathbb{R}^{D\times T}$ via a linear layer, where $D$ is the dimension of the Transformer encoder ($D$ is also typically considered as the model dimension). Instead of using deterministic, sinusoidal positional encodings, we adopt fully learnable positional encodings in our work. Our experiments showed that different positional encodings do not have a significant impact on the final performance. This could be attributed to the fact that different positional encodings still capture the relative positions within the time series. We denote the learnable positional encodings as $W_{\text{pos}}\in\mathbb{R}^{D\times T}$, and we add $W_{\text{pos}}$ to the input feature matrix $\hat{X}_{\text{enc}}$ as follows:
\begin{equation}
U_{\text{enc}} = \hat{X}_{\text{enc}} + W_{\text{pos}},    
\end{equation}
where the obtained feature matrix $U_{\text{enc}}$ will be projected as queries, keys, and values. 

\subsubsection{Input to the Teacher Branch}
Given the mask $M$ that is generated in the aforementioned encoder, we first compute the complementary mask $\overline{M}$ as follows:
\begin{equation}
    \overline{M} = \mathbbm{1} - M,
\end{equation}
where $\mathbbm{1}$ represents the matrix where all the elements equal to 1. Then, the input to the teacher branch is obtained as follows:
\begin{equation}
    \overline{X} = X \odot \overline{M}.
\end{equation}

In other words, the inputs to the encoder and the teacher branch are complementary to each other. Similarly, $\overline{X}$ is projected to the high-dimensional feature matrix $\overline{X}_{\text{tea}}\in\mathbb{R}^{D\times T}$ via a linear layer, which shares parameters with those in the encoder. The fully learnable positional encodings $W_{\text{pos}}$ are also added to the input feature matrix $\overline{X}_{\text{tea}}$ as follows:
\begin{equation}
U_{\text{tea}} = \overline{X}_{\text{tea}} + W_{\text{pos}}.
\end{equation}

\subsection{Encoder}\label{sec:enc}
The encoder $\mathcal{F}$ transforms the feature matrix $U_{\text{enc}}$ from the masked MTS input into latent representations $Z_{\text{enc}}$, while only considering the unmasked attributes.
We implement this encoder using the Transformer architecture, where $U_{\text{enc}}$ is first projected to keys, queries, and values. 
These feature matrices are then processed through a stack of Transformer blocks to generate the latent representation $Z_{\text{enc}}$.

%%%%%%%
% algorithm begin
%%%%%%%
\begin{algorithm}[t]
\caption{Procedure for Pretraining and Fine-tuning of our Proposed DMAE}\label{algorithm}
\tcc{\textbf{Pretraining}}
\KwIn{MTS training set $X_{\text{tra}}$, MTS validation set $X_{\text{val}}$, initialized model $\mathcal{M}_{0}$, number of pretraining epochs $N_{\text{tra}}$.}
\KwOut{Trained model $\mathcal{M}^{*}$.}
Generate mask $M$ based on our masking strategy; \\
$i \leftarrow 1$; \\
\For{$i\leq N_{\text{tra}}$}
{
\While{no gradient}{
Compute the supervision signal $Z_{\text{tea}}$;
}

Calculate the masked feature representations $Z_{\text{pre}}$;

Predict the masked attributes $Y$;

Calculate the loss using \eqref{eq:loss};

Update the parameters of model to get $\mathcal{M}_{i}$;

Evaluate $X_{\text{val}}$ using model $\mathcal{M}_{i}$; 
}
Select the model $\mathcal{M}^{*}$ with the lowest validation loss.

\tcc{\textbf{Fine-tuning}}
\KwIn{MTS training set $X_{\text{tra}}$, MTS validation set $X_{\text{val}}$, ground truth $\boldsymbol{y}$, trained model $\mathcal{M}^{*}$, number of fine-tuning epochs $N_{\text{tun}}$.}
\KwOut{Fine-tuned model $\mathcal{M}^{\#}$.}
Discard the teacher branch; 

Add an additional linear layer to model $\mathcal{M}^{*}$;

$i \leftarrow 1$;

\For{$i\leq N_{\text{tun}}$}
{
Predict the output $\boldsymbol{\hat{y}}$ as defined in \eqref{eq:yhat};

Calculate the task-specific loss;

Update the parameters of model  $\mathcal{M}^{*}$;

Evaluate $X_{\text{val}}$ using the updated model;
}
Select the model $\mathcal{M}^{\#}$ with the best performance.
\end{algorithm}

%%%%%%%
% algorithm end
%%%%%%%

\subsection{Teacher Branch}
The teacher branch consists of two components, one teacher encoder and one feature-query module. In the teacher encoder, the latent feature representations $Z_{\text{tea}}$ are extracted from the feature matrix $U_{\text{tea}}$ and then serve as the supervision signal for the predicted masked feature representation. The teacher encoder has the same structure and shares the parameters with the encoder defined in Sec.~\ref{sec:enc}. Notably, its gradient is stopped in the pretraining phase.

To infer latent representations $Z_{\text{pre}}$ for masked attributes from $Z_{\text{enc}}$, we introduce the feature-query module $\mathcal{H}$, constructed from Transformer blocks with cross-attention. The initial queries $Q_{\text{pre}}$, denoted as feature queries, are derived from $U_{\text{tea}}$ through projection as follows:

The feature-query module $\mathcal{H}$ is proposed to infer latent representations $Z_{\text{pre}}$ for masked attributes from $Z_{\text{enc}}$. 
This module is implemented as a stack of Transformer blocks with cross-attention.
The initial queries $Q_{\text{pre}}$, referred to as feature queries, are projected from $U_{\text{tea}}$ as follows:
\begin{equation}
    Q_{\text{pre}}=linear(U_{\text{tea}}),
\end{equation}
where the corresponding positional encodings are already incorporated in $U_{\text{tea}}$. 
The cross-attention mechanism uses identical keys and values, both derived from the latent representations $Z_{\text{enc}}$.
It is important to note that $Z_{\text{enc}}$ is fixed and not updated during this process. Finally, we can output the predicted latent feature representations $Z_{\text{pre}}$.

\subsection{Alignment Constraint}
We impose a representation alignment constraint on the predicted latent representations $\boldsymbol{Z}_{\text{pre}}$ predicted by the feature-query module, where the latent representations $\boldsymbol{Z}_{\text{tea}}$ generated from the teacher encoder serve as the supervision signals.
We then align these two feature representations by calculating the distance between them. 
In this way, the predicted representations are encouraged to lie within the encoded representation space and focus on the representation learning instead of the completion task.

\subsection{Decoder}
The decoder $\mathcal{G}$ reconstructs the masked time series attributes $Y$ from the latent feature representation $Z_{\text{pre}}$.
Unlike the vanilla transformer architecture that uses the self-attention mechanism, our decoder adopts a simple design, employing a linear layer with an activation function to decipher the extracted feature representation as follows:
\begin{equation}
    Y = \sigma(Z_{\text{pre}}; W_{\text{dec}}),
    \label{eq:dec}
\end{equation}
where $W_{\text{dec}}$ denotes the learnable parameters and $\sigma(\cdot)$ is the activation function, for which ReLU is used in our work.

\subsection{Loss Function}
The overall loss includes two terms: an alignment loss $\mathcal{L}_{\text{ali}}$ and a decoding loss $\mathcal{L}_{\text{dec}}$. The whole loss is a weighted sum as follows:
\begin{equation}
    \mathcal{L} = \mathcal{L}_{\text{dec}}(X, Y) + \lambda \mathcal{L}_{\text{ali}}(Z_{\text{tea}}, Z_{\text{pre}}),
    \label{eq:loss}
\end{equation}
where $\mathcal{L}_{\text{dec}}$ denotes the standard mean squared error (MSE) between the predicted and ground-truth masked values, and $\mathcal{L}_{\text{ali}}$ is a feature-level MSE loss that encourages the student encoder’s hidden representations $Z_{\text{pre}} \in \mathbb{R}^{D \times T}$ to match those of the teacher $Z_{\text{tea}} \in \mathbb{R}^{D \times T}$ at the same temporal positions. 

Importantly, the alignment loss $\mathcal{L}_{\text{ali}}$ is computed only on the timesteps that are masked in the student branch (i.e., those used for reconstruction). This ensures that alignment is enforced only where the model is required to infer missing information.
The hyperparameter $\lambda$ controls the relative weight of the alignment supervision. In all experiments, we set $\lambda=1$ unless otherwise specified.

\begin{table*}[!t]
    \centering
      \caption{Accuracy of the classification task on all 33 datasets from UEA Archive and UCI Machine Learning Repository. A higher accuracy indicates a better classification. The best is in bold, and the second best is underlined. ``Total Best'' denotes the count of datasets that have achieved the highest accuracy.}
      \label{tab:cls}
    \scalebox{0.8}{
  \begin{tabular}{ccccccccccccc|c}
    \toprule
    \textbf{Dataset} & Rocket & XGBoost & LSTM & DTW & SRL & TS2Vec & TNC & TS-TCC & TST & TARNet  & SVP-T  & VSFormer & \textbf{DMAE (Ours)}\\
    \midrule
    ArticularyWordRecognition 
    & \textbf{0.993} & 0.968 & 0.973 & \underline{0.987}
    & \underline{0.987} & \underline{0.987} & 0.973 & 0.953
    & 0.947 & 0.977 
    & \textbf{0.993} & \textbf{0.993} 
    &\textbf{0.993}
    \\
    
    AtrialFibrillation 
    & 0.067 & 0.267 & 0.267 & 0.220
    & 0.133 & 0.200 & 0.133 & 0.267
    & \underline{0.533} & \textbf{1.000} 
    & 0.400 & 0.467 
    & \textbf{1.000}
    \\
    
    BasicMotions 
    & \textbf{1.000} & \underline{0.976} & 0.950 & 0.975
    & \textbf{1.000} & 0.975 & 0.975 & \textbf{1.000}
    & 0.925 & \textbf{1.000} 
    & \textbf{1.000} & \textbf{1.000}
    & \textbf{1.000}
    \\
    
    CharacterTrajectories 
    & 0.991 & 0.975 & 0.985 & 0.989
    & 0.994 & \underline{0.995} & 0.967 & 0.985
    & 0.971 & 0.994 
    & 0.990 & 0.991
    & \textbf{0.997}
    \\
    
    Cricket 
    & \textbf{1.000} & 0.954 & 0.917 & \textbf{1.000}
    & \underline{0.986} & 0.972 & 0.958 & 0.917
    & 0.847 & \textbf{1.000} 
    & \textbf{1.000} & \textbf{1.000}
    & \textbf{1.000}
    \\
    
    DuckDuckGeese 
    & 0.500 & 0.695 & 0.675 & 0.600
    & 0.675 & 0.680 & 0.460 & 0.380
    & 0.300 & \textbf{0.750} 
    & \underline{0.700} & \underline{0.700}
    & \textbf{0.750}
    \\
    
    EigenWorms 
    & 0.650 & 0.539 & 0.504 & 0.618
    & \underline{0.878} & 0.847 & 0.840 & 0.779
    & 0.720 & 0.420 
    & \textbf{0.923} & 0.725
    & 0.860
    \\
    
    Epilepsy 
    & \underline{0.986} & 0.763 & 0.761 & 0.964
    & 0.957 & 0.964 & 0.957 & 0.957
    & 0.775 & \textbf{1.000} 
    & \underline{0.986} & \underline{0.986}
    & \textbf{1.000}
    \\
    
    ERing 
    & \textbf{0.989} & 0.133 & 0.133 & 0.133
    & 0.133 & 0.874 & 0.852 & 0.904
    & 0.930 & 0.919 
    & 0.937 & 0.970
    & \underline{0.974}
    \\
    
    EthanolConcentration 
    & 0.452 & 0.437 & 0.323 & 0.323
    & 0.289 & 0.308 & 0.297 & 0.285
    & 0.326 & 0.323 
    & 0.331 & \underline{0.471}
    & \textbf{0.483}
    \\
    
    FaceDetection 
    & 0.647 & 0.633 & 0.577 & 0.529
    & 0.528 & 0.501 & 0.536 & 0.544
    & \underline{0.689} & 0.641 
    & 0.512 & 0.646
    & \textbf{0.768}
    \\
    
    FingerMovements 
    & 0.520 & 0.580 & 0.580 & 0.530
    & 0.540 & 0.480 & 0.470 & 0.460
    & 0.590 & \underline{0.620} 
    & 0.600 & \textbf{0.650}
    & \underline{0.620}
    \\
    
    HandMovementDirection 
    & 0.486 & 0.478 & 0.365 & 0.231
    & 0.270 & 0.338 & 0.324 & 0.243
    & \underline{0.675} & 0.392 
    & 0.392 & 0.514
    & \textbf{0.702}
    \\
    
    Handwriting 
    & \underline{0.596} & 0.158 & 0.152 & 0.286
    & 0.533 & 0.515 & 0.249 & 0.498
    & 0.359 & 0.281 
    & 0.433 & 0.421
    & \textbf{0.608}
    \\
    
    Heartbeat 
    & 0.756 & 0.732 & 0.722 & 0.717 
    & 0.756 & 0.683 & 0.746 & 0.751
    & 0.776 & 0.780 
    & \underline{0.790} & 0.766
    & \textbf{0.822}
    \\
    
    InsectWingbeat 
    & $-$ & 0.369 & 0.176 & $-$
    & 0.160 & 0.466 & 0.469 & 0.264
    & \underline{0.687} & 0.137 
    & 0.184 & 0.200
    & \textbf{0.724}
    \\
    
    JapaneseVowels 
    & 0.962 & 0.865 & 0.797 & 0.949
    & 0.989 & 0.984 & 0.978 & 0.930
    & \textbf{0.997} & \underline{0.992} 
    & 0.978 & 0.981
    & \textbf{0.997}
    \\
 
    Libras 
    & \underline{0.906} & 0.893 & 0.856 & 0.870
    & 0.867 & 0.867 & 0.817 & 0.822
    & 0.861 & \textbf{1.000} 
    & 0.883 & 0.894
    & \textbf{1.000}
    \\

    LSST 
    & 0.635 & 0.556 & 0.373 & 0.551
    & 0.558 & 0.537 & 0.595 & 0.474
    & 0.576 & \textbf{0.976} 
    & 0.666 & 0.616
    & \underline{0.924}
    \\

    MotorImagery  
    & 0.460 & 0.459 & 0.510 & 0.500
    & 0.540 & 0.510 & 0.500 & 0.610
    & 0.610 & 0.630
    & \underline{0.650} & \underline{0.650}
    & \textbf{0.652}
    \\ 

    NATOPS 
    & 0.872 & 0.810 & 0.889 & 0.883
    & \textbf{0.944} & 0.928 & 0.911 & 0.822
    & \underline{0.939} & 0.911 
    & 0.906 & 0.933
    & \underline{0.939}
    \\

    PEMS-SF 
    & 0.751 & \textbf{0.983} & 0.399 & 0.711
    & 0.688 & 0.682 & 0.699 & 0.734
    & 0.896 & 0.936 
    & 0.867 & 0.780
    & \underline{0.979}
    \\

    PenDigits 
    & 0.981 & 0.950 & 0.978 & 0.977
    & 0.983 & \textbf{0.989} & 0.979 & 0.974
    & 0.981 & 0.976 
    & 0.983 & 0.983
    & \underline{0.987}
    \\

    Phoneme 
    & \underline{0.273} & 0.124 & 0.110 & 0.151
    & 0.246 & 0.233 & 0.207 & 0.252
    & 0.111 & 0.165 
    & 0.176 & 0.198
    & \textbf{0.298}
    \\

    RacketSports 
    & 0.901 & 0.824 & 0.803 & 0.803
    & 0.862 & 0.855 & 0.776 & 0.816
    & 0.796 & \textbf{0.987} 
    & 0.842 & \underline{0.908}
    & \textbf{0.987}
    \\

    SelfRegulationSCP1 
    & 0.908 & 0.846 & 0.689 & 0.775
    & 0.846 & 0.812 & 0.799 & 0.823
    & 0.922 & 0.816 
    & 0.884 & \underline{0.925}
    & \textbf{0.943}
    \\

    SelfRegulationSCP2 
    & 0.533 & 0.489 & 0.466 & 0.539
    & 0.556 & 0.578 & 0.550 & 0.533
    & 0.604 & 0.622
    & 0.600 & \textbf{0.644}
    & \underline{0.640}
    \\

    SpokenArabicDigits 
    & 0.712 & 0.696 & 0.319 & 0.963
    & 0.956 & \underline{0.988} & 0.934 & 0.970
    & \textbf{0.998} & 0.985 
    & 0.986 & 0.982
    & \textbf{0.998}
    \\

    StandWalkJump 
    & 0.467 & 0.333 & 0.067 & 0.200
    & 0.400 & 0.467 & 0.400 & 0.333
    & \underline{0.600} & 0.533 
    & 0.467 & 0.533
    & \textbf{0.620}
    \\

    UWaveGestureLibrary 
    & \textbf{0.944} & 0.759 & 0.412 & 0.903
    & 0.884 & 0.906 & 0.759 & 0.753
    & 0.913 & 0.878 
    & \underline{0.941} & 0.909
    & 0.937
    \\

    PAMAP2 
    & 0.931 & 0.918 & 0.949 & 0.683
    & 0.885 & 0.941 & 0.938 & 0.942
    & 0.948 & \underline{0.974} 
    & N/A & N/A 
    & \textbf{0.980}
    \\

    OpportunityGestures 
    & 0.813 & 0.755 & 0.768 & 0.762
    & 0.689 & 0.771 & 0.821 & 0.791
    & 0.732 & \underline{0.830} 
    & N/A & N/A 
    & \textbf{0.852}
    \\
    
    OpportunityLocomotion 
    & 0.875 & 0.865 & 0.900 & 0.859
    & 0.859 & 0.842 & 0.874 & 0.881
    & \underline{0.907} & \textbf{0.908} 
    & N/A & N/A
    & \textbf{0.908}
    \\
    \midrule
    \textbf{Total Best} 
    & 5 & 1 & 0 & 1 
    & 2 & 1 & 0 & 1 
    & 2 & 9
    & 4 & 5
    & \textbf{24} \\
    \textbf{Average Accuracy}
    & 0.736 & 0.660 & 0.586 & 0.662 
    & 0.684 & 0.717	& 0.689 & 0.686
    & 0.741 & 0.768
    & N/A & N/A
    & \textbf{0.847}\\
  \bottomrule
\end{tabular}
}
 % \vspace{-3mm}
\end{table*}

\subsection{Generalize to Downstream Tasks}
After the encoder has been trained through our proposed unsupervised representation learning framework, there are two widely used methods for generalizing the trained model to downstream tasks: fine-tuning and linear probing. In our work, we use the fine-tuning approach for generalizing the learned feature representations, which is putting a linear layer on top of the final feature representation and then training the whole model until it converges. It is worth noting that the teacher branch is disregarded before the fine-tuning process and the procedure for this is depicted in Alg.~\ref{algorithm}.

Formally, assume the final feature representation vector at time step $t$ is $z_{t}\in\mathbb{R}^{D}$, and the concatenated feature representation is $Z\in\mathbb{R}^{D\times T}=\left[z_{1},..., z_{t},..., z_{T}\right], t\in \left[1, T \right]$. The final representation $\boldsymbol{z}_{o}\in\mathbb{R}^{D}$ is obtained via computing the mean vector over $T$. Then, we feed $z_{o}$ into a linear layer as follows:
\begin{equation}
    \boldsymbol{\hat{y}} = W_{o} z_{o} + {b}_{o},
    \label{eq:yhat}
\end{equation}
where $W_{o}\in\mathbb{R}^{n\times D}$ and $b_{o}\in\mathbb{R}^{n\times D}$ are learnable parameters. $n$ depends on the task, i.e., the number of categories for classification, $n=1$ for regression, or the prediction period length for forecasting.

\section{Experiments}
We thoroughly assess the effectiveness of DMAE through extensive experiments on three downstream tasks, namely classification, regression, and forecasting.

\subsection{Experimental Setup}
\subsubsection{Classification} We use the benchmark time-series datasets UEA Archive~\cite{bagnall2018uea} and UCI Machine Learning Repository~\cite{asuncion2007uci}, in which a total of 33 datasets. We compare our proposed method DMAE with the following 10 existing methods: ROCKET~\cite{dempster2020rocket}, XGBoost~\cite{chen2016xgboost}, LSTM~\cite{hochreiter1997long}, DTW~\cite{lei2017similarity}, SRL~\cite{franceschi2019unsupervised}, 
TS2Vec~\cite{yue2022ts2vec},
TNC~\cite{tonekaboni2021unsupervised},
TS-TCC~\cite{eldele2021time},
TST~\cite{zerveas2021transformer}, TARNet~\cite{chowdhury2022tarnet}, SVP-T~\cite{zuo2023svp} and VSFormer~\cite{xi2024vsformer}. Classification accuracy is used as the metric.
    
\subsubsection{Regression}  For regression, 6 datasets~\cite{tan2020monash} are used: AppliancesEnergy, BenzeneConcentr, BeijingPM10, BeijingPM25, LiveFuelMoisture, and IEEEPPG. These 6 datasets cover a wide range of attribute dimensionality, series length, and sample number. We compare our method with 11 baselines that are reported in ~\cite{tan2020monash} and we also compare with methods TST~\cite{zerveas2021transformer} and TARNet~\cite{chowdhury2022tarnet}. The Root Mean Squared Error (RMSE) is used as the regression metric.

\begin{table*}[!t]
\centering
  \caption{RMSE of the regression task on 6 datasets. A smaller RMSE indicates a better regression. The best is in bold, and the second best is underlined. ``Total Best'' denotes the count of datasets that have achieved the lowest RMSE.}
  \label{tab:reg}
  \scalebox{0.98}{
  \begin{tabular}{ccccccc|ccc}
    \toprule
    \textbf{Method}
    & Appliances. & Benzene. & BeijingPM10.
    & BeijingPM25. & LiveFuel. & IEEEPPG. & \textbf{Total Best} & \textbf{Average}\\
    \midrule
    SVR
    & 3.457 & 4.790 & 110.574 
    & 75.734 & 43.021 & 36.301 & 0 &45.646\\
    
    RandomForest
    & 3.455 & 0.855 & 94.072 
    & 63.301 & 44.657 & 32.109 & 0 & 39.741 \\
 
    XGBoost
    & 3.489 & 0.637 & 93.138 
    & 59.495 & 44.295 & 31.487 & 0 & 38.757\\
    
    1-NN-ED
    & 5.231 & 6.535 & 139.229 
    & 88.193 & 58.238 & 33.208 & 0 & 55.106\\
    
    5-NN-ED
    & 4.227 & 5.844 & 115.669 
    & 74.156 & 46.331 & 27.111 & 0 & 45.556\\
    
    1-NNDTW
    & 6.036 & 4.983 & 139.134 
    & 88.256 & 57.111 & 37.140 & 0 & 55.443\\
    
    5-NNDTW
    & 4.019 & 4.868 & 115.502 
    & 72.717 & 46.290 & 33.572 & 0 & 46.161\\
    
    Rocket
    & 2.299 & 3.360 & 120.057 
    & 62.769 & 41.829 & 36.515 & 0 & 44.472\\
    
    FCN
    & 2.865 & 4.988 & 94.348
    & 59.726 & 47.877 & 34.325 & 0 & 40.688\\
    
    ResNet
    & 3.065 & 4.061 & 95.489 
    & 64.462 & 51.632 & 33.150 & 0 & 41.977\\
    
    Inception
    & 4.435 & 1.584 & 96.749 
    & 62.227 & 51.539 & \underline{23.903} & 0 & 40.073\\
    
    TST
    & 2.375 & 0.494 & \underline{86.866} 
    & \underline{53.492} & 43.138 & 27.806 & 0 & \underline{35.695}\\

    TARNet
    & \underline{2.173} & \underline{0.481} & 90.482 & 60.271 & \textbf{41.091} & 26.372 & \underline{1} & 36.812\\

    \midrule
    \textbf{DMAE (Ours)}
    & \textbf{1.866} & \textbf{0.353} & \textbf{82.239} 
    & \textbf{48.124} & \underline{41.688} & \textbf{18.841} & \textbf{5} & \textbf{32.185}\\
  \bottomrule
\end{tabular}
}
\end{table*}

\begin{table*}[!t]
\centering
  \caption{MSE/MAE of the forecasting task on 4 datasets. All results are averaged from 4 different choices of prediction period $O \in \{96, 192, 336, 720\}$. A smaller MSE or MAE indicates better forecasting. The best is in bold, and the second best is underlined.}
  \label{tab:fore}
  \scalebox{0.98}{
  \begin{tabular}{cccccccc|c}
    \toprule
    \textbf{Method} 
    & ETTh1 & ETTh2 & ETTm1 & ETTm2 & Weather & Electricity & Traffic & \textbf{Average}\\
    \midrule
    TS2Vec &
    0.446/0.456 &
    0.417/0.468 &
    0.699/0.557 &
    0.326/0.361 &
    0.233/0.267 &
    0.213/0.293 &
    0.470/0.350 &
    0.401/0.393 \\
    
    CoST &
    0.485/0.472 &
    0.399/0.427 &
    0.356/0.385 &
    0.314/0.365 &
    0.324/0.329 &
    0.215/0.295 &
    0.435/0.362 &
    0.361/0.376 \\
    
    TF-C &
    0.637/0.638 &
    0.398/0.398 &
    0.744/0.652 &
    1.755/0.947 &
    0.286/0.349 &
    0.363/0.398 &
    0.717/0.456 &
    0.700/0.548 \\
    
    LaST &
    0.474/0.461 &
    0.499/0.497 &
    0.398/0.398 &
    0.255/0.326 &
    0.232/0.261 &
    0.186/0.274 &
    0.713/0.397 &
    0.394/0.373 \\

    TST  &
    0.466/0.462 &
    0.404/0.421 &
    0.373/0.389 &
    0.297/0.347 &
    0.239/0.276 &
    0.209/0.289 &
    0.586/0.362 &
    0.368/0.364 \\
    
    TimesNet &
    0.458/0.450 &
    0.414/0.427 &
    0.400/0.406 &
    0.291/0.333 &
    0.259/0.287 &
    0.192/0.295 &
    0.620/0.336 & 
    0.376/0.362 \\
    
    Ti-MAE &
    0.423/0.446 &
    0.423/0.446 &
    0.366/0.391 &
    0.264/0.320 &
    0.230/\underline{0.265} &
    0.205/0.296 &
    0.475/0.310 &
    0.335/0.345 \\
    
    InfoTS &
    0.451/0.471 &  
    0.429/0.468 &  
    0.398/0.417 &  
    0.285/0.345 &  
    0.258/0.291 &  
    0.217/0.332 &  
    0.495/0.336 &
    0.362/0.380 \\
    
    SimMTM  &
    \underline{0.404}/\underline{0.428} &
    \underline{0.348}/\underline{0.391} &
    \textbf{0.340}/\underline{0.379} &
    \underline{0.260}/\underline{0.318} &
    \underline{0.228}/0.267 &
    \underline{0.162}/\underline{0.256} &
    \underline{0.392}/\underline{0.264} &
    \underline{0.305}/\underline{0.329} \\

    SimTS &
    0.426/0.452 &
    0.368/0.425 &
    0.359/0.403 &
    0.279/0.347 &
    0.243/0.296 &
    0.188/0.283 &
    0.408/0.298 &
    0.324/0.358 \\
    
    \midrule
    \textbf{DMAE (Ours)} &
    \textbf{0.382}/\textbf{0.402} &
    \textbf{0.335}/\textbf{0.372} &
    \underline{0.346}/\textbf{0.375} &
    \textbf{0.232}/\textbf{0.294} &
    \textbf{0.214}/\textbf{0.256} &
    \textbf{0.144}/\textbf{0.231} &
    \textbf{0.348}/\textbf{0.236} &
    \textbf{0.286}/\textbf{0.309} \\
  \bottomrule
\end{tabular}
}
\end{table*}

\subsubsection{Forecasting} We have followed the standard experimental setups in baseline~\cite{dong2023simmtm} where used 4 datasets: ETT~\cite{zhou2021informer} that contains 4 subsets, Weather~\footnote{https://www.bgc-jena.mpg.de/wetter/}, Traffic~\footnote{http://pems.dot.ca.gov}, and Electricit\footnote{https://archive.ics.uci.edu/ml/datasets/}. 
We compare our method with 8 existing state-of-the-art baselines: TS2Vec~\cite{yue2022ts2vec}, CoST~\cite{woo2021cost}, TF-C~\cite{zhang2022self}, LaST~\cite{wang2022learning}, TST~\cite{zerveas2021transformer}, TimesNet~\cite{wu2022timesnet}, Ti-MAE~\cite{li2023ti}, SimMTM~\cite{dong2023simmtm},
InfoTS~\cite{luo2023time},
and SimTS~\cite{zheng2024simple}.
We use both MSE and MAE as the metrics.

\subsection{Implementation Details}
In our transformer-based encoder, we stack 3 transformer blocks with 16 attention heads in each. We set the feature dimension $D=128$ for input $U$ and set the dimension of feature matrices $Q$, $K$, and $V$ as 256. As for the unsupervised training, we train our model for a total of 500 epochs with an initial learning rate of 0.001. We set $\lambda =1$ to control the balance of the two loss terms. For fine-tuning, we train our model for a total of 200 epochs with an initial learning rate of 0.001. We use Adam~\cite{kingma2014adam} as the optimizer, and our whole models are implemented using Pytorch~\cite{paszke2019pytorch}.
All datasets used in our experiments are publicly available and correctly referenced.
Our code will be released upon publication, including scripts for training, evaluation, and reproducing all reported results and visualizations.

\subsection{Main Results}
\subsubsection{Classification}
The results presented in Tab.~\ref{tab:cls} illustrate that our DMAE method achieves the highest classification accuracy in 24 out of 33 datasets, with an impressive average accuracy of 0.847. This performance surpasses the TARNet method by a margin of 0.079.
For the latest baseline methods, SVP-T and VSFormer, they achieve the highest classification accuracy on 4 and 5 out of 33 datasets, respectively. In comparison, our method significantly outperforms both.
Given the extensive variety of datasets and comparative baselines, it is highly unlikely for any single model to dominate across all datasets, as noted in~\cite{chowdhury2022tarnet}. Despite the varying scales and dimensions across these datasets, our DMAE method consistently ranks as either the best or the second-best in terms of accuracy. This consistency demonstrates the robustness and general adaptability of our pretext task designs, affirming their efficacy across a wide spectrum of data characteristics.

\subsubsection{Regression}
The regression results (RMSE) are shown in Tab.~\ref{tab:reg}. It can be observed that our proposed method achieves the best performance on 5 out of 6 datasets. Overall, the average RMSE of our method is $32.185$, which achieves a $9.8\%$ RMSE reduction compared to the previous SOTA method TST. The results also validate the effectiveness of our proposed method for the downstream regression task. 

\subsubsection{Forecasting}
Tab.~\ref{tab:fore} demonstrates the forecasting results MSE and MAE on 4 datasets. The forecasting task requires the method to predict the future $O$ time points, where $O \in \{96, 192, 336, 720\}$. We report the average MSE/MAE results of these 4 choices. It can be observed that our proposed method consistently outperforms other pretraining methods on all the datasets except the subset ETTm1. The average MSE/MAE of our method is $0.286/0.309$, which achieves a $6.23\%$ MSE reduction and $6.07\%$ MAE reduction compared to the previous SOTA method SimMTM. The results verify the effectiveness of our proposed method for the downstream forecasting task. 

\subsection{Ablation Study}
\subsubsection{Normalization}
We adopt batch normalization in our method to replace layer normalization because batch normalization is more robust than layer normalization when handling the outlier issue. Tab.~\ref{tab:abl} demonstrates the performance differences of using different normalizations on the three downstream tasks. It can be observed that using batch normalization consistently outperforms using layer normalization, which empirically verifies the robustness of using batch normalization in the time series domain.

\subsubsection{Dual Masking}
As introduced in the main paper, our model is trained with two terms of losses as follows:
\begin{equation}
    \mathcal{L} = \mathcal{L}_{\text{dec}}(X, Y) + \lambda \mathcal{L}_{\text{ali}}(\boldsymbol{Z}_{\text{tea}}, \boldsymbol{Z}_{\text{pre}}).
\end{equation}

To assess the impact of our proposed pretext tasks on model performance, we conduct an ablation study focusing on the individual contributions of $\mathcal{L}_{\text{ali}}$ and $\mathcal{L}_{\text{dec}}$ to the training process. Specifically, we trained our model exclusively on $\mathcal{L}_{\text{ali}}$ and $\mathcal{L}_{\text{dec}}$ to evaluate their separate effects. The results of this investigation are shown in Tab.~\ref{tab:abl:loss}. In scenarios where DMAE is trained without $\mathcal{L}_{\text{ali}}$, we eliminate the alignment loss from the training process, effectively transforming our model into a Transformer-based architecture that prioritizes reconstruction tasks. Conversely, when training DMAE without $\mathcal{L}_{\text{dec}}$, the decoding loss is omitted, thereby limiting the model's function to estimating masked features without attempting to reconstruct the absent attributes.

The results demonstrate that employing the full loss framework yields superior performance across all evaluated downstream tasks. By comparing the performance of DMAE in the absence of $\mathcal{L}_{\text{ali}}$ to the complete DMAE model, it becomes evident that our masked feature estimation approach significantly benefits time series representation learning, enhancing the model's capacity for learning representations. Furthermore, the comparison between DMAE without $\mathcal{L}_{\text{dec}}$ and the complete DMAE model highlights the importance of predicting masked attributes within time series data. Such predictions facilitate the model's understanding of temporal relationships between attributes,
reinforcing the necessity of both $\mathcal{L}_{\text{ali}}$ and $\mathcal{L}_{\text{dec}}$ in the training objective.

\subsubsection{Masking Strategy}
We investigate the choice of using another masking strategy: independent masking with a Bernoulli distribution. We set the sequence length $T=60$ and the number of variables $M=20$. We set $r=0.15$ in span masking and $p=r=0.15$ in independent masking. An illustration of these two masking strategies can be found in the appendix, and it is obvious that there are more continuous attributes masked in span masking than the independent masking. The empirical results of using independent masking are also shown in Tab.~\ref{tab:abl}. It can be observed that the performance of using the independent masking strategy is consistently worse than using the span masking strategy via geometric distribution. The possible reason is that the span masking strategy has a certain probability of masking consecutive time stamps. In this way, the temporal relations of time series are learned during the completion process, thus enhancing the representation learning capacity.

\subsubsection{Masking Ratio}
Additionally, we study the hyperparameter of the masking ratio $r$. We set different ratios $r \in \{0.1, 0.15, 0.2, 0.25, 0.5\}$ and conduct experiments as show in Tab.~\ref{tab:ratio}. We can observe that the optimal performance is achieved when $r=0.15$. However, when $r=0.5$, there is a noticeable decrease in performance for both tasks. This is attributed to the increased level of missing information, which exacerbates the difficulty of the reconstruction process.

\begin{table*}[!t]
\centering
  \caption{Ablation study of our DMAE method on 3 downstream tasks, and we select 3 datasets for each task. $^{\#}$ indicates using layer normalization and $^{*}$ indicates using the independent masking strategy.}
  \label{tab:abl}
  \scalebox{0.98}{
  \begin{tabular}{cccc|ccc|ccc}
    \toprule
    \textbf{Method} & Heart. & Insect. & PEMS-SF. & Benzene. & BeijingPM25. & LiveFuel. & ETTh1 & Weather & Traffic\\
    \midrule
    \textbf{DMAE$^{\#}$} & 0.753 & 0.684 & 0.928 & 0.945 & 57.332 & 44.223 & 0.487/0.481 & 0.276/0.305 & 0.453/0.382\\
    
    \textbf{DMAE$^{*}$} & 0.783  & 0.667  & 0.942 & 0.432  & 54.331  & 42.876  & 0.412/0.446 & 0.232/0.261 & 0.386/0.273 \\
    
    \textbf{DMAE}  & \textbf{0.822} & \textbf{0.724} & \textbf{0.979} & \textbf{0.353} & \textbf{48.124} & \textbf{41.688} & \textbf{0.382}/\textbf{0.402} & \textbf{0.214}/\textbf{0.256} &  \textbf{0.348}/\textbf{0.236}\\
  \bottomrule
\end{tabular}
}
 % \vspace{-3mm}
\end{table*}

\begin{table*}[!t]
\centering
  \caption{Ablation study of our DMAE method on 3 downstream tasks and we select 3 datasets for each task.}
  \label{tab:abl:loss}
  \scalebox{0.98}{
  \begin{tabular}{c|ccc|ccc|ccc}
    \toprule
    \multirow{2}{*}{Variant} & \multicolumn{3}{|c|}{Classification (Accuracy)} 
    & \multicolumn{3}{|c|}{Regression (RMSE)}
    & \multicolumn{3}{c}{Forecasting (MSE/MAE)}\\
    \cmidrule(lr){2-10}
    & Heart. & Insect. & PEMS-SF. & Benzene. & BeijingPM25. & LiveFuel. & ETTh1 & Weather & Electricity \\
    \midrule  
    DMAE w/o $\mathcal{L}_{\text{ali}}$
    & 0.785 & 0.642 & 0.864
    & 0.526 & 56.221 & 46.784
    & 0.526/0.621 & 0.183/0.498 & 0.211/0.288
    \\
    DMAE w/o $\mathcal{L}_{\text{dec}}$
    & 0.764 & 0.657 & 0.892
    & 0.639 & 58.764 & 45.354
    & 0.662/0.704 & 0.243/0.466 & 0.243/0.302
    \\

    DMAE &  \textbf{0.822} & \textbf{0.724} & \textbf{0.979} & \textbf{0.353} & \textbf{48.124} & \textbf{41.688} & \textbf{0.382}/\textbf{0.402} & \textbf{0.214}/\textbf{0.256} & \textbf{0.144}/\textbf{0.231} \\
  \bottomrule
\end{tabular}
}
\end{table*}

\begin{table}[!t]
\centering
  \caption{Masking ratio study of our DMAE method on the classification and regression task, 3 datasets are selected for each task.}
  \label{tab:ratio}
  \scalebox{0.8}{
  \begin{tabular}{lccc|ccc}
    \toprule
    \textbf{Ratio} & Heart. & Insect. & PEMS-SF. & Benzene. & BeijingPM25. & LiveFuel. \\
    \midrule  
    $r=0.1$ & 0.810 & 0.712 & \textbf{0.979}&  0.411 & 48.369 & 41.956\\
    $r=0.15$ &  \textbf{0.822} & \textbf{0.724} & \textbf{0.979} & \textbf{0.353} & \textbf{48.124} & \textbf{41.688} \\
    $r=0.2$ & 0.818 & 0.709 & 0.968 & 0.382 & 48.211 & 42.024 \\
    $r=0.25$& 0.802 & 0.710 & 0.968 & 0.394 & 48.337 & 42.218 \\
    $r=0.5$ & 0.760 & 0.684 & 0.902 & 0.435 & 49.120 & 43.462\\
  \bottomrule
\end{tabular}
}
\end{table}

\begin{figure}[!t]
\centering
\includegraphics[width=0.95\linewidth]{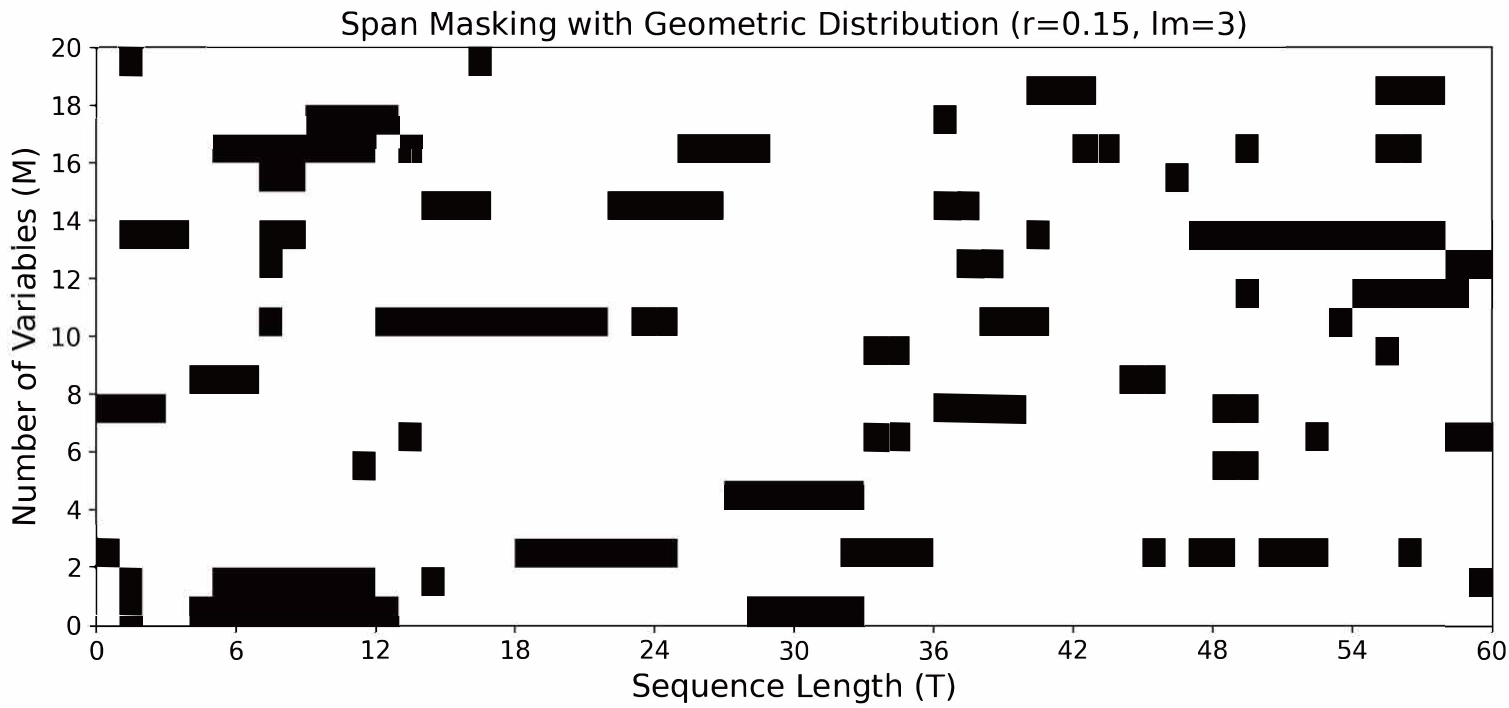} \\
\includegraphics[width=0.95\linewidth]{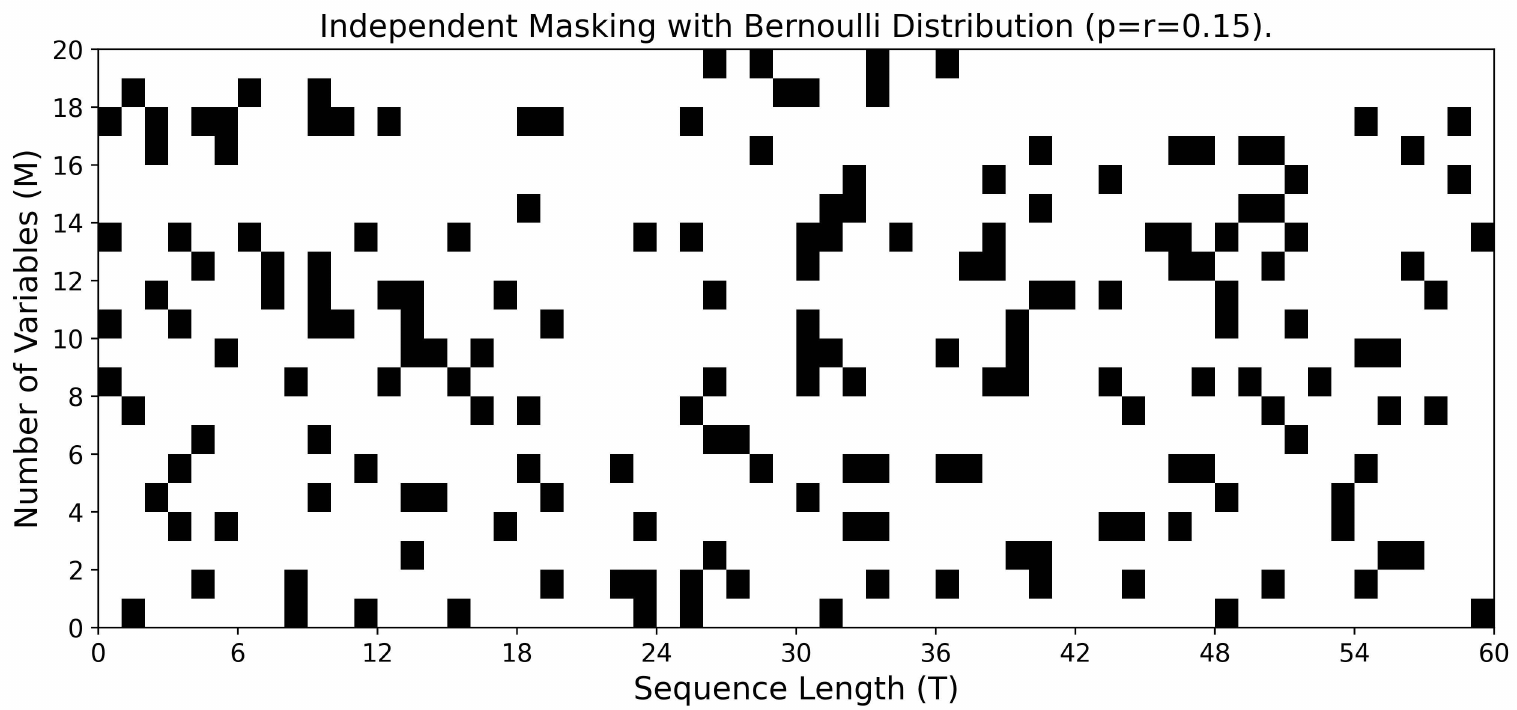}
\caption{Examples of two different masking strategies, where the above is the span masking used in our work and the below is the independent masking. The small black square indicates this attribute is masked.}
\label{fig:masking}
\end{figure}

\subsubsection{Training Efficiency}
Although DMAE introduces an additional teacher branch and alignment loss during training, the actual overhead is modest. The student and teacher encoders share weights, and the teacher is only used for latent supervision during training. At inference time, only the student encoder and decoder are retained, leading to the same runtime latency as a standard autoencoder-based baseline.

We further profile the training overhead introduced by the alignment loss across different downstream tasks. Specifically, enabling $\mathcal{L}_{\text{ali}}$ increases total training time by approximately 4.2\% for classification, 3.5\% for regression, and 5.1\% for forecasting tasks. This small cost is consistently offset by substantial gains in representation quality.

\subsection{Qualitative Results}
\subsubsection{Visualization of Alignment Effect}
To better understand the effect of the alignment loss $\mathcal{L}_{\text{ali}}$, we visualize the cosine similarity between the predicted latent features $Z_{\text{pre}}$ and the teacher-provided targets $Z_{\text{tea}}$ across timesteps. We evaluate models trained with and without the alignment loss on the Heartbeat classification dataset.

\begin{figure}[t]
    \centering  \includegraphics[width=\linewidth]{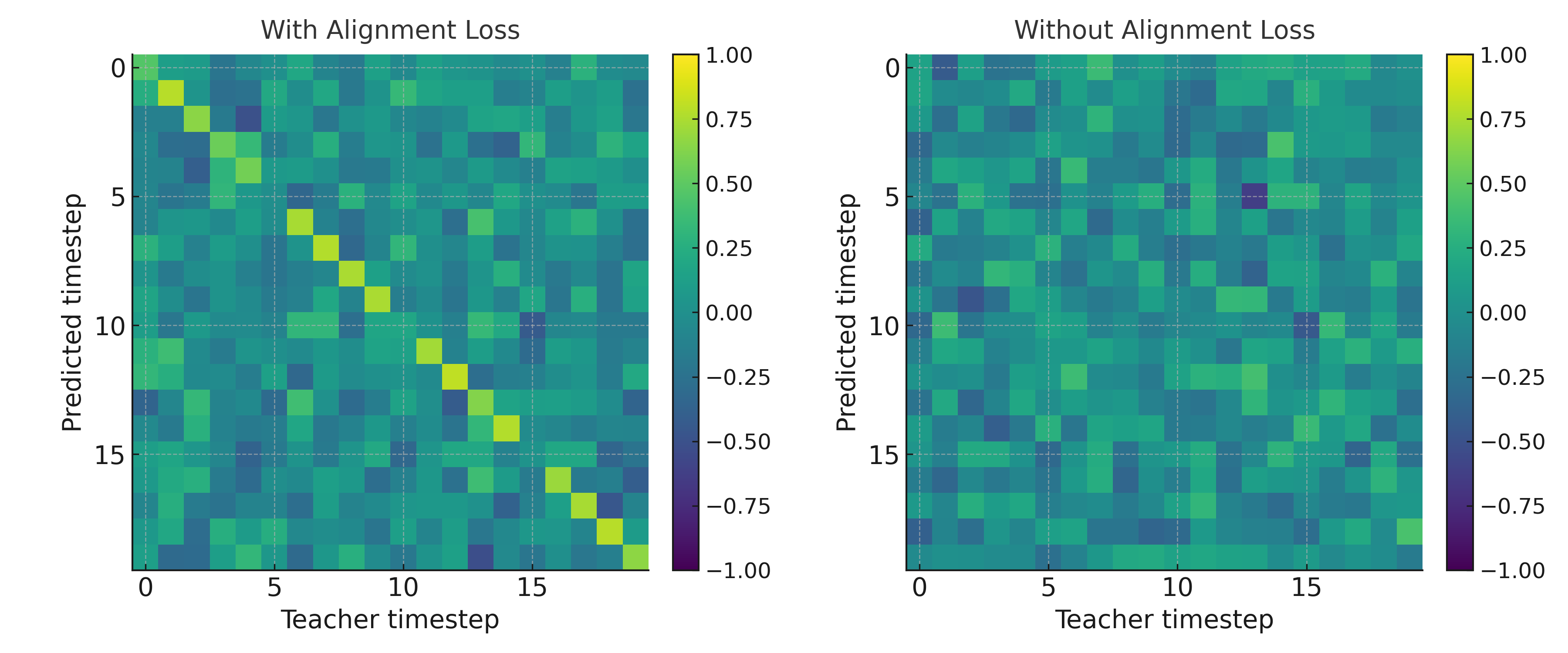}
    \caption{Cosine similarity between $Z_{\text{pre}}$ and $Z_{\text{tea}}$ across timesteps. The model trained with alignment loss (left) shows clearer diagonal patterns than the one without (right), indicating better feature-level correspondence.}
    \label{fig:zpre_ztea_alignment}
\end{figure}

Fig.~\ref{fig:zpre_ztea_alignment} shows cosine similarity heatmaps between $Z_{\text{pre}} \in \mathbb{R}^{D \times T}$ and $Z_{\text{tea}} \in \mathbb{R}^{D \times T}$ for a randomly selected test sequence. With alignment loss, the heatmap exhibits a clearer diagonal tendency, suggesting the stronger timestep-wise correspondence between student and teacher representations. Without alignment, the similarity is more diffuse and lacks temporal alignment.

These visualizations confirm that the alignment loss promotes consistency in the latent space, leading to better-structured and semantically aligned feature representations.

\begin{figure}[t]
    \centering
    \includegraphics[width=0.98\linewidth]{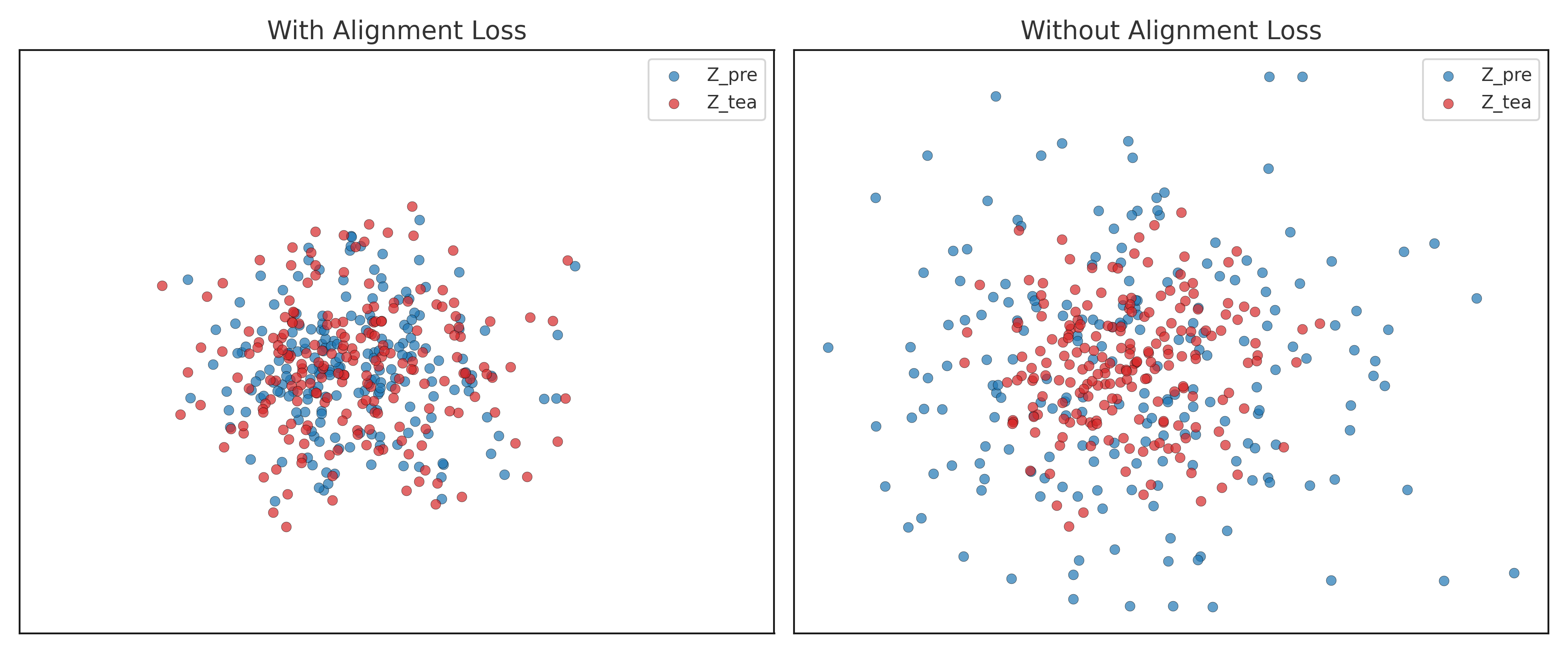}
    \caption{
    PCA visualization of predicted features $Z_{\text{pre}}$ and teacher features $Z_{\text{tea}}$. 
    With alignment loss (left), $Z_{\text{pre}}$ and $Z_{\text{tea}}$ show strong distributional overlap. 
    Without alignment (right), the predicted features drift noticeably from the teacher features, 
    indicating degraded alignment in the latent space.
    }
    \label{fig:pca_alignment}
\end{figure}

To further examine how the alignment loss $\mathcal{L}_{\text{ali}}$ impacts the distribution of learned features, we visualize the predicted features $Z_{\text{pre}}$ and teacher features $Z_{\text{tea}}$ using PCA~\cite{abdi2010principal}. As shown in Fig.~\ref{fig:pca_alignment}, the model trained with alignment loss produces $Z_{\text{pre}}$ features that closely align with the distribution of $Z_{\text{tea}}$, forming overlapping clusters in the PCA space. In contrast, without alignment loss, the predicted features exhibit noticeable drift from the teacher features, indicating reduced consistency in the latent space. This visualization highlights the effectiveness of $\mathcal{L}_{\text{ali}}$ in preserving semantic consistency and stabilizing the feature learning process.

\begin{figure}[t]
    \centering
    \includegraphics[width=0.48\linewidth]{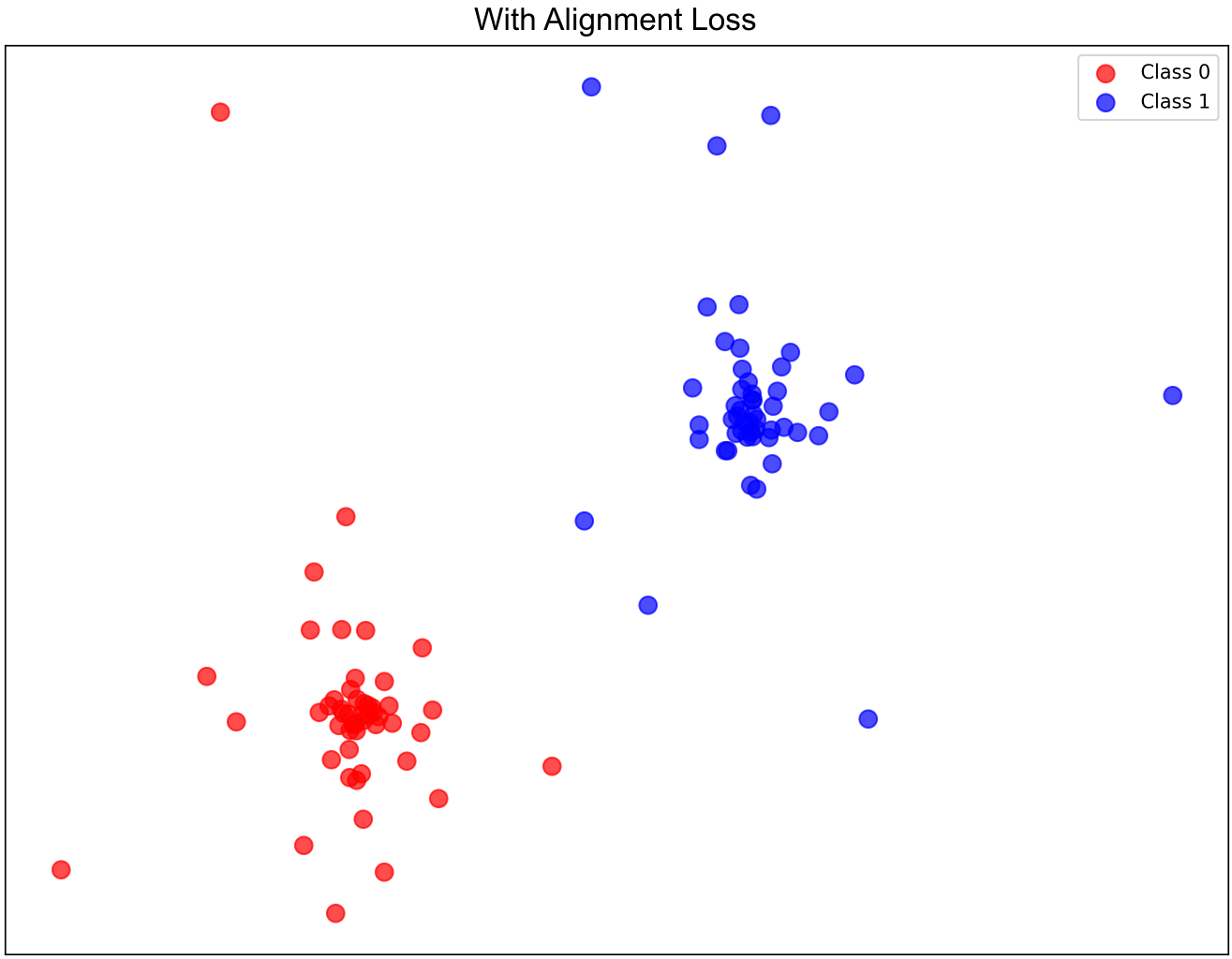}
    \hfill
\includegraphics[width=0.48\linewidth]{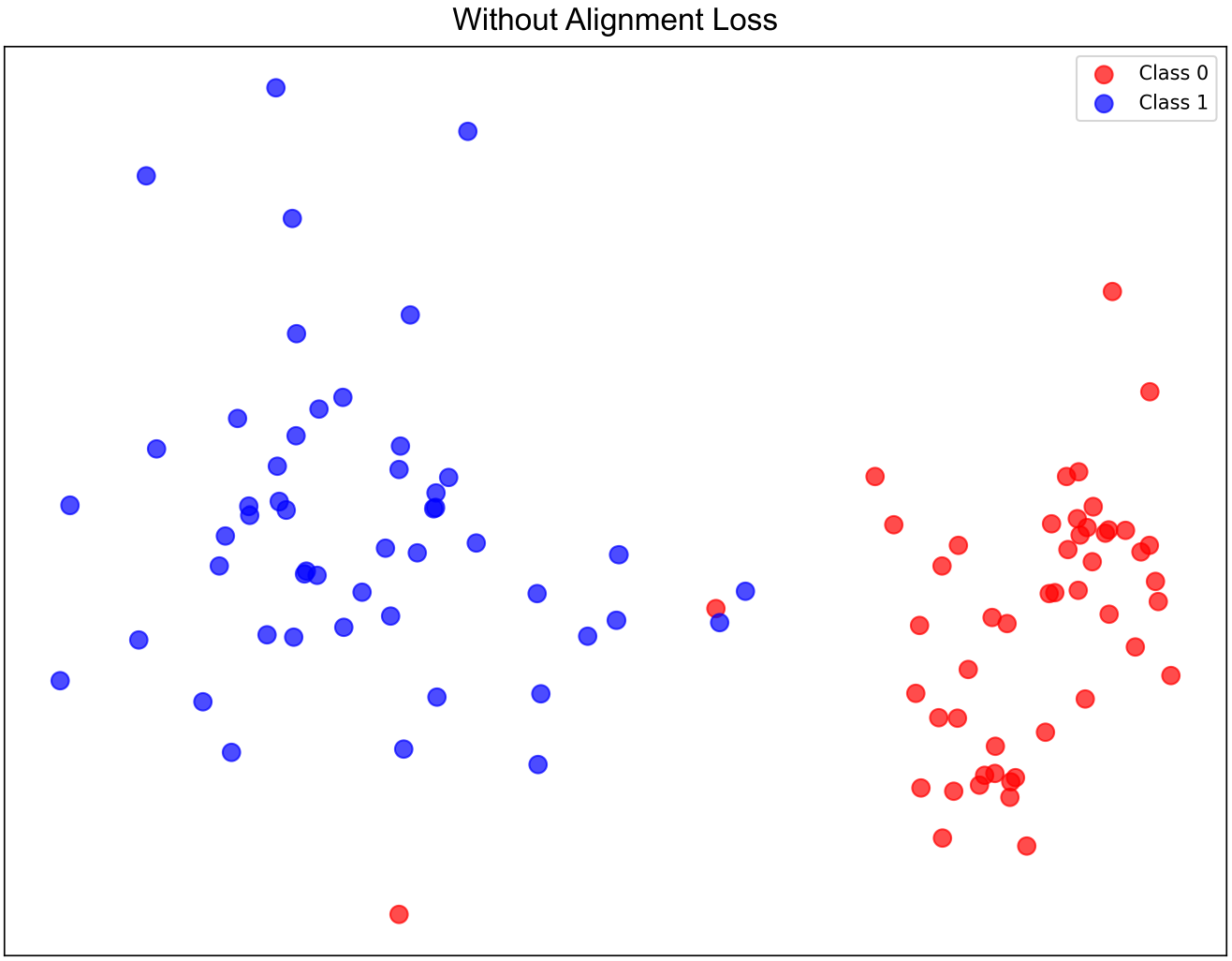}
    \caption{t-SNE visualization of the predicted features $Z_{\text{pre}}$ under aligned and unaligned settings.}
\label{fig:tsne_alignment}
\end{figure}

To better understand how the alignment loss $\mathcal{L}_{\text{ali}}$ shapes the latent space, we visualize the predicted features $Z_{\text{pre}}$ using t-SNE. As shown in Fig.~\ref{fig:tsne_alignment}, when the alignment loss is applied, the class clusters become more compact and exhibit moderate overlap, indicating improved structure in the learned representation. In contrast, removing the alignment loss causes the clusters to drift and partially collapse into each other, leading to noisier feature distributions. These results highlight the role of $\mathcal{L}_{\text{ali}}$ in encouraging semantically consistent and distinguishable feature spaces.

\section{Motivations and Insights}
While reconstructing masked features is not new in computer vision~\cite{chen2021empirical,li2023scaling}, our work is among the first to successfully apply this strategy to multivariate time series representation learning. Below, we summarize the key motivations behind our dual completion design:

\begin{itemize}
    \item \textbf{Global-local feature learning.} Reconstructing masked values focuses on local temporal fidelity, while reconstructing masked latent features encourages the model to capture global structure and inter-feature dependencies. Combining both objectives enables the model to balance fine-grained detail recovery with high-level semantic understanding.
    \item \textbf{Improved robustness and representation balance.} The dual masking mechanism reduces over-reliance on dominant features by forcing the model to attend to multiple dimensions and mitigate feature bias. This also increases robustness to noise and missing data, which are common in real-world time series.
    \item \textbf{Broader task compatibility.} Time series tasks often involve complex cross-temporal and cross-dimensional relationships. Our dual-task design encourages learning representations that generalize well across forecasting, classification, and regression tasks.
\end{itemize}
Together, these motivations lead to a richer and more structured representation space, enhancing generalization and performance across diverse datasets. The consistent empirical improvements observed in three downstream tasks support the effectiveness of our approach.

\section{Conclusion}
We present DMAE, a novel Dual-Masked Autoencoder framework for unsupervised multivariate time series representation learning. DMAE introduces two complementary pretext tasks, masked value reconstruction and latent feature estimation, integrated via a shared encoder and a teacher-student alignment mechanism. By enforcing a feature-level alignment constraint, DMAE promotes temporal consistency and semantic structure in the latent space.
Extensive experiments across classification, regression, and forecasting tasks demonstrate that DMAE consistently outperforms state-of-the-art baselines. The learned representations are generalizable and transferable across datasets and tasks. Despite its dual-branch training design, DMAE remains computationally efficient due to parameter sharing and the exclusion of the teacher branch during inference.

\noindent\textbf{Limitations and Future Work.}
One limitation lies in the reliance on the Transformer backbone, which may not be optimal for all types of time series data.
In future work, we aim to explore alternative temporal architectures and develop lightweight variants of DMAE. We also plan to enhance domain generalization by enabling robust single-round pretraining across heterogeneous datasets and tasks.

\section*{Acknowledgment}
This material is based upon work supported by the Air Force Office of Scientific Research under award number FA9550-23-1-0290.

\bibliographystyle{IEEEtran}
\bibliography{IEEEexample}

\end{document}